\documentclass[10pt,twocolumn,letterpaper]{article}

\usepackage{cvpr}              

\usepackage{bbm}
\usepackage{graphicx}
\usepackage{amsmath}
\usepackage{amssymb}
\usepackage{amsthm}
\usepackage{mathrsfs}
\usepackage{amsfonts}
\usepackage{booktabs}
\usepackage{multirow}
\usepackage{multicol}
\usepackage[table]{xcolor}
\usepackage{color, colortbl}
\usepackage{pifont}
\newcommand{\cmark}{\ding{51}}%

\usepackage[pagebackref,breaklinks,colorlinks]{hyperref}

\usepackage[capitalize]{cleveref}
\crefname{section}{Sec.}{Secs.}
\Crefname{section}{Section}{Sections}
\Crefname{table}{Table}{Tables}
\crefname{table}{Tab.}{Tabs.}

\def\eg{\emph{e.g.}} 
\def\ie{\emph{i.e.}} 

\def\etc{\emph{etc.}}

\begin{document}

\title{Perceive, Interact, Predict:  Learning Dynamic and Static Clues \\for End-to-End Motion Prediction}

\author{Bo Jiang\textsuperscript{1,2*},\quad Shaoyu Chen\textsuperscript{1,2*},\quad Xinggang Wang\textsuperscript{1 \textsuperscript{$\dagger$}},\quad Bencheng Liao\textsuperscript{1,2}, \quad
Tianheng Cheng\textsuperscript{1,2} \\  Jiajie Chen\textsuperscript{2},\quad Helong Zhou\textsuperscript{2},\quad Qian Zhang\textsuperscript{2},\quad Wenyu Liu\textsuperscript{1}, \quad Chang Huang\textsuperscript{2}\\
\textsuperscript{1} Huazhong University of Science \& Technology,\quad
\textsuperscript{2} Horizon Robotics\\
{\tt\small \{bjiang, shaoyuchen, xgwang, bcliao, thch, liuwy\}@hust.edu.cn}\\
{\tt\small \{jiajie.chen, helong.zhou, qian01.zhang, chang.huang\}@horizon.ai}\\
}
\maketitle

\let\thefootnote\relax\footnote{*: Equal contribution.}
\let\thefootnote\relax\footnote{$\dagger$: Corresponding author.}

\begin{abstract}
Motion prediction is highly relevant to the perception of dynamic objects and static map elements in the scenarios of autonomous driving. 
In this work, we propose PIP, the first  end-to-end Transformer-based framework which jointly and interactively performs online mapping, object detection and motion prediction. 
PIP leverages map queries, agent queries and mode queries to  encode the instance-wise information of map elements, agents and motion intentions, respectively. 
Based on the unified query representation, a differentiable multi-task interaction scheme is proposed to exploit the correlation between perception and prediction. 
Even without human-annotated HD map or agent's historical tracking trajectory as guidance information, PIP realizes
end-to-end multi-agent motion prediction and 
achieves better performance than tracking-based and HD-map-based methods. 
PIP provides comprehensive high-level information of the driving scene (vectorized static map and dynamic objects with motion information), and contributes to the downstream planning and control.
Code and models will be released for facilitating further research.
\end{abstract}

\section{Introduction}
\label{sec:intro}
Motion prediction aims at forecasting the future trajectory of  other agents (road participants like vehicles) in the driving scene,  which provides prior guidance on robust planning and control of autonomous vehicles and avoids collision.
Intuitively, accurate motion prediction relies on a comprehensive understanding of the surrounding environment. On one hand, motion prediction requires the position information of target agents. Accurate detection is the basis for inferring the motion intention of  agents. On the other hand, static map can provide effective prior information for forecasting the future behavior of dynamic objects (\eg, vehicles usually drive along the road lane). 
Therefore, motion prediction is highly correlated with the perception of the moving objects and static map elements.

\begin{figure}[]
\centering
\includegraphics[width=0.47\textwidth]{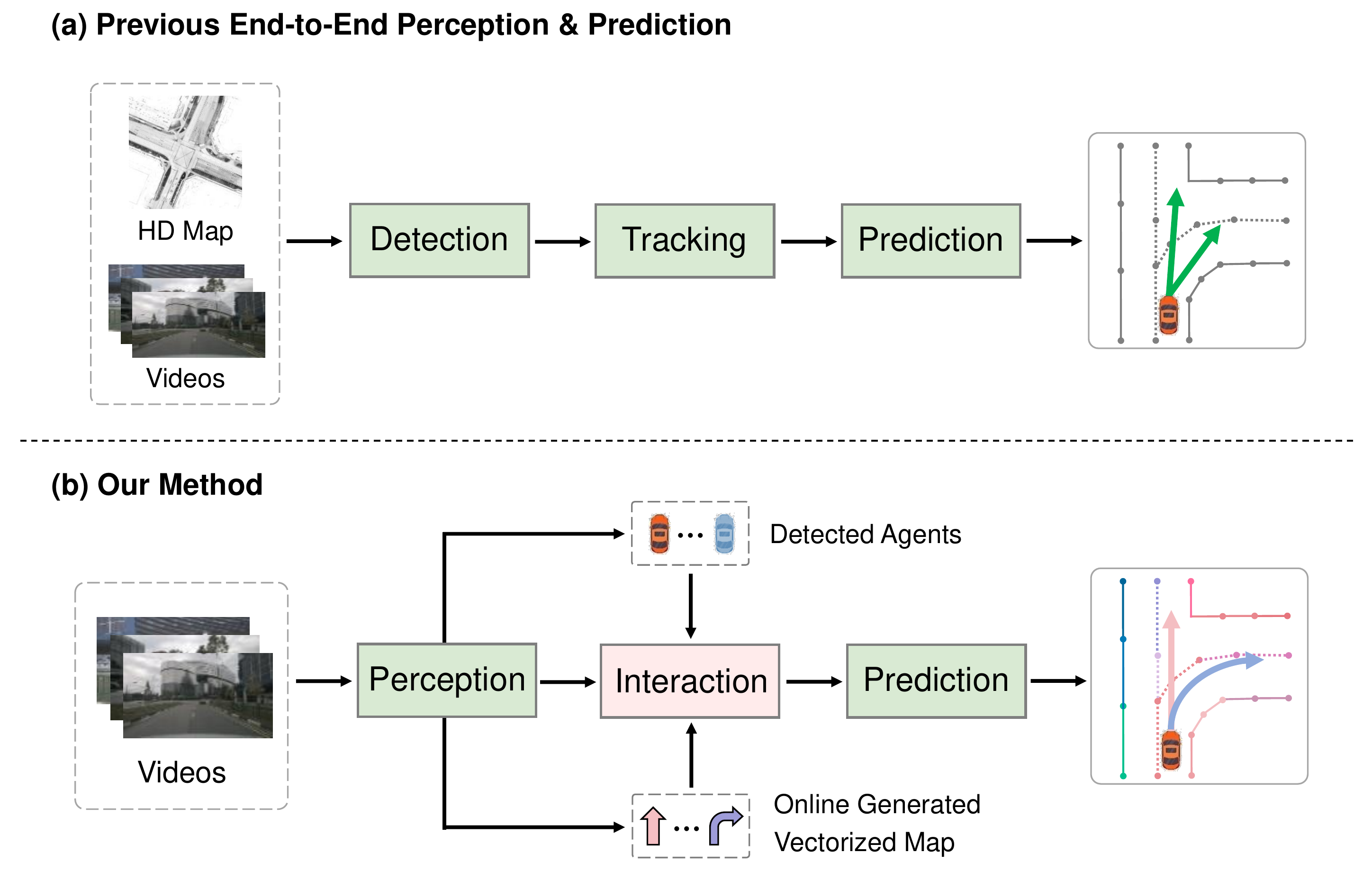}
\caption{Comparison of end-to-end motion prediction paradigms. (a) Previous perception and prediction paradigm requires human-annotated HD map which greatly limits its application. (b) Our proposed method PIP only uses video input, and generates vectorized map in an online fashion. PIP extracts dynamic and static clues from videos as prior guidance, and adopts a multi-task interaction scheme for better motion prediction.}
\vspace{-1mm}
\label{fig:intro}
\end{figure}

Traditional motion prediction methods\cite{chai2019multipath, gao2020vectornet, liang2020lanegcn, liu2021mmtrans} decouple perception from prediction.
They take perception results and offline-constructed high-definition (HD) map as input for motion prediction.
Since  perception and prediction are fully detached, the prediction module cannot directly learn from sensor data and miss raw information. And the accumulated error in the perception module also affects the prediction.

Recent works\cite{luo2018fnf, casas2018intentnet, djuric2021multixnet, liang2020pnpnet, gu2022vip3d} focus on end-to-end motion prediction (Figure~\ref{fig:intro}), \ie,  taking sensor data and HD map as input, instead of perception results and HD map. 
Some works\cite{liang2020pnpnet, gu2022vip3d} adopt a detection-tracking-prediction pipeline. Some methods\cite{luo2018fnf, casas2018intentnet, djuric2021multixnet} predict motion with hand-designed post-processing steps. 
All the existing works rely on offline generated HD map, which introduces many problems (high cost, restriction of law, freshness of map, \etc) and  limits the real-scene applications of these methods.

In this work, we propose PIP (\textbf{P}erceive, \textbf{I}nteract, \textbf{P}redict), the first  end-to-end Transformer-based framework which jointly and interactively performs online mapping, object detection and motion prediction.
PIP directly learns dynamic and static clues from vehicle-mounted cameras, not requiring human-annotated HD map.

PIP leverages map queries, agent queries and  mode queries to respectively encode the instance-wise information of map elements, agents and motion intentions.
Based on the unified query representation, a tailored Motion Interactor is proposed to exploit the high correlation between perception and motion prediction.
Specifically, to reduce the variance among agents, 
we perform agent-wise normalization by switching from ego-centric coordinate system to agent-centric coordinate system. 
Besides, for each agent, we filter map information according to confidence score and spatial relation, and encode the information both explicitly and implicitly for effective motion-map interaction.
PIP is fully differentiable and end-to-end trained,  not requiring hand-designed post-processing steps.

Though without human-annotated HD map or agent's historical trajectory as guidance, PIP achieves better motion prediction performance than HD-map-based and tracking-based methods. PIP achieves 0.258 EPA, which is higher than the previous state-of-the-art result\cite{gu2022vip3d}. Meanwhile, PIP greatly reduces minADE and minFDE by 0.91m and 1.17m.
Besides, thorough ablation experiments demonstrate the effectiveness of the interaction design.

The contribution can be summarized as follows:
\begin{itemize}
\item PIP is the first end-to-end Transformer-based framework which jointly and interactively performs online mapping, object detection and motion prediction. PIP directly learns dynamic and static clues from vehicle-mounted cameras, not requiring human-annotated HD map or agent's historical trajectory. 

\item We design a multi-task interaction scheme to exploit the correlation between perception and prediction. PIP is fully differentiable and end-to-end trained, without hand-designed post-processing steps.

\item Even without HD map or agent's historical trajectory as guidance information, PIP achieves better motion prediction performance than tracking-based and HD-map-based methods. PIP can provide comprehensive high-level information of the driving scene (vectorized static map and dynamic objects with motion information), which are vital for the downstream planning and control.
\end{itemize}

\begin{figure*}[h]
\centering
\vspace{1mm}
\includegraphics[width=0.98\textwidth]{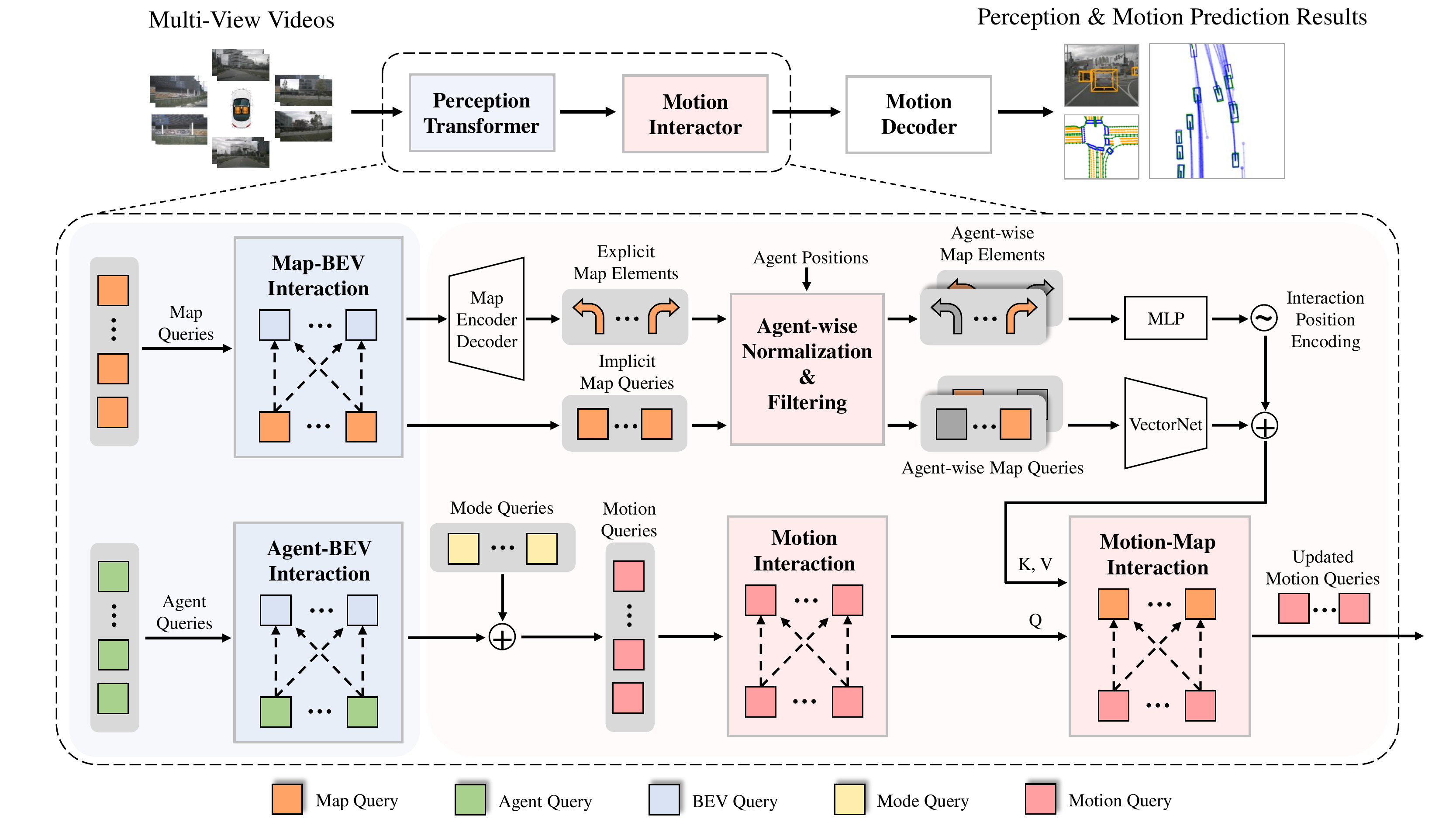} 
\caption{Overall architecture of PIP. Perception Transformer learns static and dynamic clues from multi-view videos with map queries and agent queries. 
Motion queries interact with agent-wise map information both explicitly and implicitly, and then output motion prediction.}

\label{fig:framework}
\end{figure*}

\section{Related Work}
\label{sec:related}
\paragraph{3D Object Detection.} LiDAR-based 3D object detection methods\cite{zhou2018voxelnet, lang2019pointpillars, yin2021centerpoint} have the best performance among all sensor modalities thanks to LiDAR's accurate depth estimation. But recent camera-based methods\cite{wang2021fcos3d, wang2022detr3d, li2022bevformer} have also made remarkable progress. Compared with LiDAR point clouds, camera images naturally contain more diverse scene information, and can capture objects at longer distance. FCOS3D\cite{wang2021fcos3d} is a one-stage detector for monocular 3D detection, inspired from the classical 2D detector FCOS\cite{tian2019fcos}. DETR3D\cite{wang2022detr3d} uses learnable 3D queries to learn features from 2D images, and makes prediction in an end-to-end fashion. Recent works\cite{liu2022petrv2, li2022bevformer, chen2022gkt, liu2022vision} transform image features into bird’s-eye-view (BEV), which provides a unified feature representation for downstream tasks such as detection and segmentation. Based on these methods, PIP adopts BEV queries and agent queries to perform image-to-BEV feature transformation and 3D object detection, and further extends the framework to online mapping and motion prediction.

\paragraph{Online Mapping.} Map is essential for autonomous driving. Early works\cite{philion2020lss, hu2021fiery, zhou2022cvt} regard map construction as a segmentation task in image space or BEV space. Recent works \cite{li2022hdmapnet, liu2022vectormapnet, liao2022maptr} focus vectorized map representation, which contains instance-level semantic information of road topology. HDMapNet\cite{li2022hdmapnet} predicts lane segmentation and other lane attributes in BEV space, and constructs vectorized map with post-processing steps. VectorMapNet\cite{liu2022vectormapnet} treats map elements as vector polylines, and predicts map elements in a regressive way. MapTR\cite{liao2022maptr} leverages map queries to learn features from BEV feature map, and directly predicts all vectorized map elements simultaneously. PIP uses map queries for learning static scene information, inspired by MapTR\cite{liao2022maptr}.
PIP further encodes map features both implicitly and explicitly and performs effective interaction between online mapping and motion prediction.

\paragraph{Traditional Motion Prediction.} Traditional motion prediction methods\cite{chai2019multipath, gao2020vectornet, liang2020lanegcn, liu2021mmtrans} take the output of the perception module as input, including historical detection and tracking results and human-annotated HD map. Some works\cite{chai2019multipath, phan2020covernet} convert input into rasterized scene images in BEV, and use CNN-based models to solve the problem. Some other works\cite{gao2020vectornet, liang2020lanegcn, liu2021mmtrans} use vectorized input, and adopt graph models or attention-based models to learn and predict. In terms of trajectory decoding, regression-based models\cite{gao2020vectornet, liu2021mmtrans} directly predict future trajectory. Goal-based models\cite{zhao2020tnt, gu2021densetnt} first predict trajectory endpoints, then complete the full trajectory based on the predicted endpoints. Heatmap-based models\cite{gilles2021home, gilles2022gohome} predict future probability distributions of endpoints instead of predicting coordinates.

\paragraph{End-to-End Motion Prediction.} End-to-End methods directly use sensor data inputs to perform perception and prediction. Some works\cite{hu2021fiery, zhang2022beverse} predict future motions at scene-level, FIERY\cite{hu2021fiery} predicts per-grid occupancy and future forward flow in BEV space. BEVerse\cite{zhang2022beverse} proposes an iterative flow to generate future states more efficiently. In this paper, we focus on agent-level end-to-end motion prediction task. FaF\cite{luo2018fnf} is an early attempt, but its prediction horizon is very limited (only 30 ms). IntentNet\cite{casas2018intentnet} decouples motion prediction into intention prediction and trajectory completion. FaF\cite{luo2018fnf} and IntentNet\cite{casas2018intentnet} both rely on post-processing steps to generate complete trajectory, which are time-consuming and inefficient. PnPNet\cite{liang2020pnpnet} uses historical trajectories and local BEV features around each agent to predict future motions. All of the above works\cite{luo2018fnf, casas2018intentnet, liang2020pnpnet} rely on LiDAR data and HD map. Recently proposed method ViP3D\cite{gu2022vip3d} uses camera images and HD map as input, and follows MOTR\cite{zeng2021motr} to complete detection and tracking via agent queries. HD map will be used to provide static scene information for motion prediction. HD map introduces many problems (high cost, restriction of law, freshness of map, \etc), and limits the real-scene applications of these methods. In contrast, PIP only relies on camera input, getting rid of human-annotated HD map. PIP takes full advantage of the high correlation between perception and motion prediction, and utilizes the rich perceptual features as effective prior information to boost the performance of motion prediction.  

\section{Method}
\label{sec:method}

\subsection{Overall Architecture}
\label{sec:arch}
The overall architecture of PIP is depicted in Figure \ref{fig:framework}, First, PIP takes multi-view videos as input, and uses a backbone network (e.g., ResNet-50\cite{he2016resnet}) to extract image features in each frame, then the multi-view video feature maps are fed into a BEV encoder to transform the temporal 2D features to BEV space (\S \ref{sec:encoder}). Second, we leverage two sets of queries (map queries and agent queries) to learn static and dynamic features from BEV feature maps respectively(\S \ref{sec:encoder}). Third, we adopt a small set of mode queries to represent different motion intentions and capture intention-aware scene features, and combine them with the agent queries to form our motion queries. Then we propose the Motion Interactor to exploit the correlation between perception and prediction(\S \ref{sec:interactor}) via attention mechanism. Next, we introduce the motion decoder to generate motion prediction output from the motion queries (\S \ref{sec:decoder}). Finally, we utilize a multi-task loss to train the network (\S \ref{sec:loss}).

\subsection{Perception Transformer}
\label{sec:encoder}
\paragraph{BEV Encoder.} PIP leverages a BEV encoder to transform  multi-view  images to the unified BEV representation. Given multi-view video images $I = \{I^i_t \in \mathbb{R}^{H_I \times W_I \times 3}\}_{i=1, t=1}^{N_{v}, T_h}$, where ${T_h}$ is the number of historical frames and ${N_{v}}$ is the number of camera views, a backbone network is used to generate multi-view video feature maps $F = \{F^i_t \in \mathbb{R}^{H_F \times W_F \times C}\}_{i=1, t=1}^{N_{v}, T_h}$, then we utilize a group of BEV queries $Q_{\rm BEV} \in \mathbb{R}^{H_Q \times W_Q \times C}$~\cite{li2022bevformer,zhou2022cvt} for sampling features from $F$ and complete the transformation from perspective view to BEV view.

\paragraph{Map Encoder \& Decoder.} We adopt a group of map queries $Q_{M} \in \mathbb{R}^{N_{I} \times N_{P} \times C}$ to complete online mapping~\cite{liao2022maptr}, where $N_{I}$ denotes map instance number and $N_{P}$ denotes point number of each map instance. Each map instance is represented as an ordered set of map points. The map encoder outputs the updated map queries. Then map decoder will take them as inputs, and predicts map instance class scores $\hat{c}_{M}$ and locations of each map point $p_{M}$ via a classification branch and a regression branch, respectively.

\vspace{-2mm}
\paragraph{Detection Encoder \& Decoder.} We adopt a group of agent queries $Q_{A} \in \mathbb{R}^{N_{A} \times C}$to learn temporal motion features of dynamic agents from BEV queries $Q_{\rm BEV}$~\cite{li2022bevformer}. Each detection encoder layer consists of multi-head self-attention (MHSA) among agent queries, deformable attention~\cite{zhu2020deformable} between agent queries and BEV queries, and a feed forward layer. The detection decoder includes a classification branch and a regression branch. The classification branch outputs the detection score, and the regression branch outputs the location $p_{A}$ and other attributes of the predicted agents.

\begin{figure}[h]
\centering
\includegraphics[width=0.45\textwidth]{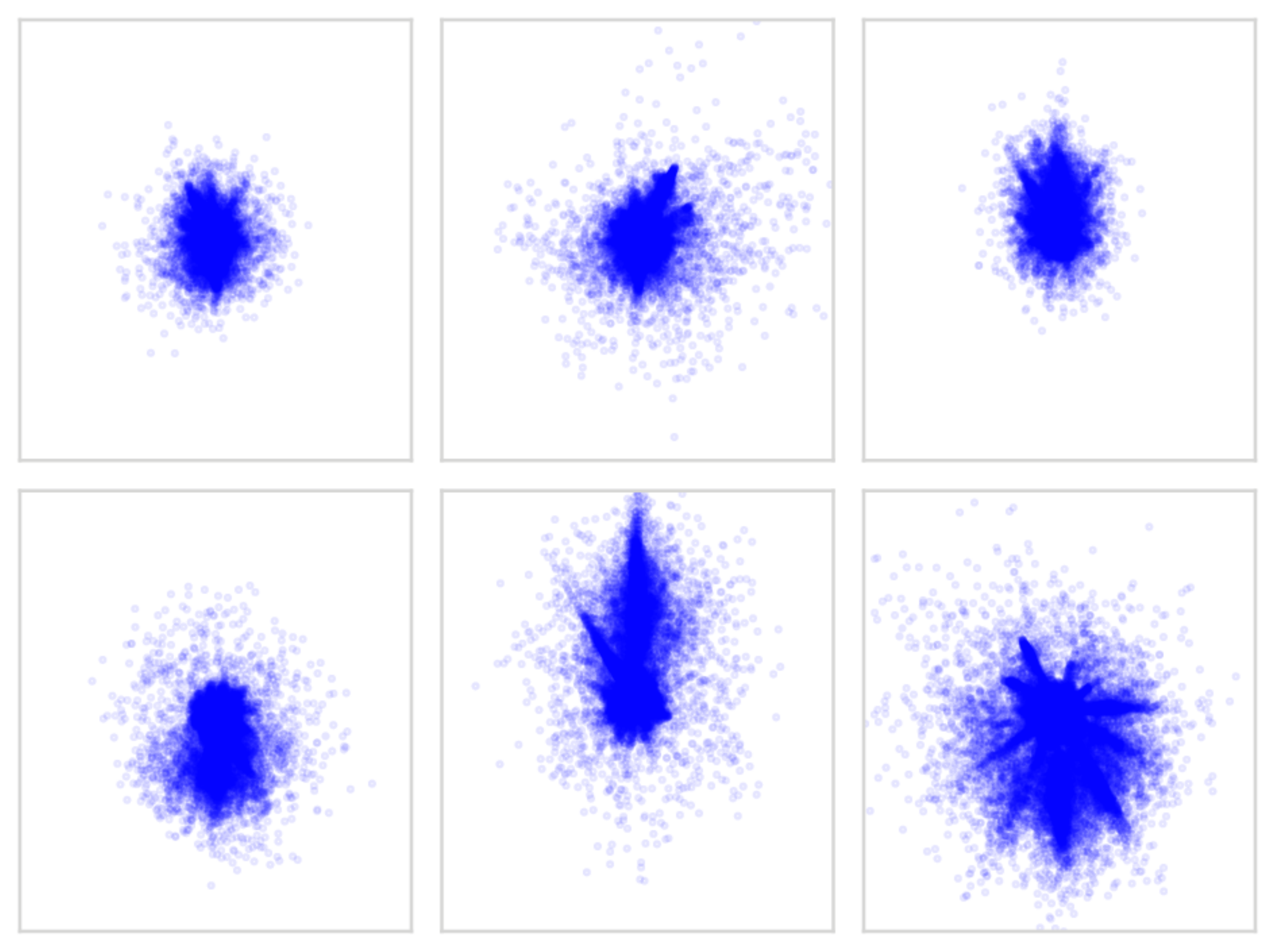}
\caption{Visualization of predicted endpoints of different mode queries on the nuScenes\cite{caesar2020nuscenes} val set. For each prediction, we use the mean value of different mode trajectory endpoints as the origin, and use the mean orientation of different trajectory at the final timestamp as the positive direction of y-axis.}
\label{fig:mode}
\end{figure}

\subsection{Motion Interactor}
\label{sec:interactor}

\paragraph{Motion Encoding.}
In real-world driving scenarios, drivers need to have a comprehensive understanding of the environment, and meanwhile focus on some certain areas based on their motion intentions. In order to achieve accurate multi-modal motion prediction, we propose a small group of mode queries $Q_{\rm mode} \in \mathbb{R}^{N_{\rm mode} \times C}$ initialized with orthogonal values to represent different motion intentions for each dynamic agent. By combining mode queries $Q_{\rm mode}$ and agent queries $Q_{A}$, the \textit{j}-th future motion mode of the \textit{i}-th agent can be formulated as follows:

\begin{equation}
\label{eq:motion}
    Q^{ij}_{\rm motion} = Q^{i}_{A} + Q^{j}_{\rm mode}
\end{equation}

$Q_{\rm motion} \in \mathbb{R}^{N_{A} \times N_{\rm mode} \times C}$ denotes the motion queries. Each motion query will be used for motion prediction of its corresponding agent and mode. Figure \ref{fig:mode} is the visualization of predicted endpoints of different mode queries on the nuScenes\cite{caesar2020nuscenes} val set. It's obvious that different mode queries focus on different motion intentions. For example, the mode query corresponding to the top middle figure is more likely to make right-turn prediction, and the mode query corresponding to the bottom middle figure makes more long straight predictions.

Then we feed the motion queries into a self-attention module for interaction among different motion intentions as well as for interaction among different agents. The output is the updated motion queries which contain dynamic information of the scene.

\begin{equation}
\label{eq:motion-self-attn}
    Q^{\rm SA}_{\rm motion} = {\rm FFN(Self \text{-} Attention}(Q_{\rm motion}))
\end{equation}

Next, we want to leverage map information for motion prediction. We discover that directly feeding the vectorized output of online mapping module into the motion prediction module is suboptimal.

We further process the map information for more effective interaction between online mapping and motion prediction.

\paragraph{Agent-wise Normalization.}
Unlike traditional motion prediction task which only considers one target agent, we perform motion prediction for all agents in the scene simultaneously.  Agents may appear anywhere within the perception range. 
In the ego-centric coordinate system, two agents at different locations may have very similar future trajectory patterns (e.g., go straight). But their future trajectory coordinates vary greatly, as well as the map coordinates.

To reduce the variance among agents, we perform coordinate
normalization for each agent. Specifically, ground-truth future trajectories are transformed from absolute coordinates to per-step offsets, so that the regression range is reduced. Besides, we transform coordinates of map from original ego-centric coordinate system to the agent-centric coordinate system: $\hat{p}^{ij}_{M} = p^{i}_{M} - p^{j}_{A}$.

\paragraph{Agent-wise Filtering.}
Previous works take ground-truth HD map around the target agent as input, while we use predicted map. Predicted map elements with low confidence scores are likely to be false-positive predictions and provide wrong guidance. Besides, the scene range is quite large (102.4m $\times$ 102.4m) compared with previous works\cite{liu2021mmtrans, liu2022vectormapnet}.  Actually, for each agent, distant map elements provide little guidance for motion prediction, and may disturb them from focusing on the key map elements.

Therefore, for each agent, we further filter the predicted map elements based on a confidence threshold $\tau$ and a distance threshold $\mu$. The valid map elements for interaction with the \textit{j}-th agent are formulated as:

\begin{equation}
\label{eq:filter}
    \hat{Q}^{j}_{M} = \{q_i:q_i \in Q_M, {\rm Scr}(q_i) \ge \tau, {\rm Dis}(q_i, j) \le \mu \}
\end{equation}

\noindent where $\hat{Q}^{j}_{M} \in \mathbb{R}^{N_{I}^{j} \times N_P \times C}$ and $q_i \in \mathbb{R}^{N_{P} \times C}$. ${\rm Scr}(q_i) = \max(\hat{c}_{M}^{i})$ is the max classification score of map query $q_i$ across all classes decoded by the map head. ${\rm Dis}(q_i, j) = \underset{pt \in q_i}{\rm min}||pt -p_{A}^{j}||_2$ is the distance between the closet point of $q_i$ and the \textit{j}-th agent.

\paragraph{Map Encoding.}
The vanilla map queries are point-level features, missing map connectivity information and map element relationships. we adopt a VectorNet\cite{gao2020vectornet} to further extract map features from point-level to instance-level, obtaining map instance queries. For the \textit{j}-th agent, the map instance queries are formulated as:

\begin{equation}
\label{eq:vectornet}
    Q_{I}^{j} = {\rm VectorNet}(\hat{Q}_{M}^{j}), \, Q_{I}^{j} \in \mathbb{R}^{N_{I}^{j} \times C}
\end{equation}

Besides, we leverage the closet point of each map instance from a target agent as the representative point:

\begin{equation}
\label{eq:closest-point}
    p^{ij}_{I} = \hat{p}^{i\hat{k}}_{M}, \hat{k} = \underset{k \in N_{P}}{{\arg\min}} \, ||\hat{p}^{ik}_{M} - p^{j}_{A}||_{2}
\end{equation}

$p^{ij}_{I}$ denotes the representative point of the \textit{i}-th map instance for the \textit{j}-th agent. $p^{ik}_{M}$ denotes the \textit{k}-th point of the \textit{i}-th map instance.
Then we use these points to generate interaction position encoding for each map instance by a single MLP: ${\rm PE}(p^{ij}_{I}) = {\rm MLP}_{\rm PE}(p^{ij}_{I})$. The interaction position encoding provides explicit map position information, which helps motion queries to know where to look and focus.

\paragraph{Motion-Map Interaction.}
Finally, we perform interaction between motion queries and the refined map queries via a cross-attention module:

\begin{equation}
\begin{aligned}
\label{eq:interactor}
    &Q^{\rm CA}_{\rm motion} = {\rm FFN(Cross \text{-} Attention}(Q, K, V, K_{\rm pos}) \\
    &Q = Q^{\rm SA}_{\rm motion}, \, K = V = Q_I, \, K_{\rm pos} = {\rm PE}(p_{I})
\end{aligned}
\end{equation}

The final updated motion queries $\hat{Q}_{\rm motion}$ are the concatenation of $Q^{\rm SA}_{\rm motion}$ and $Q^{\rm CA}_{\rm motion}$ in the channel dimension:

\begin{equation}
\label{eq:update-motion}
    \hat{Q}_{\rm motion} = {\rm Concat(Q^{\rm SA}_{\rm motion}, Q^{\rm CA}_{\rm motion})}
\end{equation}

\subsection{Motion Decoder}
\label{sec:decoder}

The motion decoder takes the updated motion queries as input, and predicts trajectories in per-step offset format using a regression branch, then we can get the full trajectories $\hat{S} \in \mathbb{R}^{N_{A} \times T_f \times 2}$ by accumulating the per-step offsets during inference. $T_f$ is the number of future frames. PIP is also compatible with other motion decoding strategies such as goal-based\cite{zhao2020tnt, gu2021densetnt} or heatmap-based\cite{gilles2021home, gilles2022gohome} methods, but we just adopt regression-based decoder head for its simplicity and effectiveness.

\subsection{Loss}
\label{sec:loss}
\begin{table*}[]
\begin{center}
\renewcommand{\tabcolsep}{3.0pt}
\resizebox{0.7\textwidth}{!}{
\begin{tabular}{l|c|c|c c c c}
\midrule
\multirow{2}{*}{Method} & Trajectory & Requiring  & \multirow{2}{*}{EPA$\uparrow$} & \multirow{2}{*}{minADE$\downarrow$} & \multirow{2}{*}{minFDE$\downarrow$} & \multirow{2}{*}{MR$\downarrow$} \\
& Decoding & HD Map & & & & \\
\midrule
Traditional & Goal & \cmark  & 0.233 & 2.64 & 4.37 & 0.286 \\
PnPNet~\cite{liang2020pnpnet} & Goal &\cmark & 0.250 & 2.73 & 4.59 & 0.252 \\
ViP3D~\cite{gu2022vip3d} & Goal &\cmark & 0.250 & 2.52 & 4.05 & 0.242 \\
\midrule
Traditional & Heatmap &\cmark  & 0.243 & 2.72 & 4.16 & 0.268 \\
PnPNet~\cite{liang2020pnpnet} & Heatmap &\cmark & 0.249 & 2.73 & 4.17 & 0.255 \\
ViP3D~\cite{gu2022vip3d} & Heatmap &\cmark & 0.268 & 2.62 & 3.94 & 0.220 \\
\midrule
Traditional & Regression &\cmark  & 0.234 & 2.26 & 3.22 & 0.284 \\
PnPNet~\cite{liang2020pnpnet} & Regression &\cmark & 0.244 & 2.13 & 2.96 & 0.258 \\
ViP3D~\cite{gu2022vip3d} & Regression &\cmark & 0.253 & 2.14 & 2.92 & 0.254 \\
\midrule
\multicolumn{1}{c|}{PIP} & Regression & HD Map Free & 0.258 & 1.23 & 1.75 & 0.195 \\
\bottomrule
\end{tabular}}
\end{center}
\vspace{-2mm}
\caption{End-to-end motion prediction results on the nuScenes \cite{caesar2020nuscenes} val set. "Goal", "Heatmap" and "Regression" indicate goal-based\cite{zhao2020tnt}, heatmap-based\cite{gilles2021home} and regression based\cite{cui2019mtp} trajectory decoding methods, respectively.}
\label{tbl:mainresults}
\end{table*}

\paragraph{Online Mapping.} The loss for online mapping is composed of a instance classification loss and a point regression loss. First we have a Focal\cite{lin2017focal} loss for classification between predicted map elements $\hat{c}^{\hat\omega(i)}_{M}$ and the assigned class labels $c^{i}_{M}$ according to the optimal map instance matching result ${\hat\omega}$:

\begin{equation}
\label{eq:loss_map_cls}
    \mathcal{L}_{{\rm map\_cls}} = \sum_{i=0}^{N_{I}-1} \mathcal{L}_{{\rm Focal}}(\hat{c}^{\hat\omega(i)}_{M}, c^{i}_{M})
\end{equation}
After the instance-level matching, we do a point-level matching between each predicted and ground-truth map instance pair, and then we calculate the point regression loss as a Manhattan distance $\mathbb{D}_{\rm M}$ between the predicted point ${\hat{p}_{\hat\omega(i),j}}$ and its corresponding ground-truth point $p_{i,\hat\gamma_i(j)}$ according to the optimal point-level matching result ${\hat\gamma}$:

\begin{equation}
\label{eq:loss_map_reg}
    \mathcal{L}_{{\rm map\_reg}} = \sum_{i=0}^{N_{I}-1} \mathbbm{1}_{\{c^{i}_{M} \ne \varnothing\}} \sum_{j=0}^{N_{P}-1} \mathbb{D}_{\rm M}({\hat{p}_{\hat\omega(i),j}}, p_{i,\hat\gamma_i(j)})
\end{equation}

\paragraph{Object Detection.} For each predicted agent, we will assign a class label $c^{i}_{A}$ to it (or $\varnothing$ label for 'no object') according to the optimal agent assignment result ${\hat\sigma}$ between predicted and ground-truth set. $\hat{c}^{\hat\sigma(i)}_{A}$ is the  predicted classification score of the $i$-th agent in ${\hat\sigma}$, then the classification loss for object detection is a Focal\cite{lin2017focal} loss formulated as:

\begin{equation}
\label{eq:loss_det_cls}
    \mathcal{L}_{{\rm det\_cls}} = \sum_{i=0}^{N_{A}-1} \mathcal{L}_{{\rm Focal}}(\hat{c}^{\hat\sigma(i)}_{A}, c^{i}_{A})
\end{equation}

The regression loss is the $\textit{l}_1$ loss between the predicted boxes $\hat{b}^{\hat\sigma}$ and the ground-truth boxes $b$:

\begin{equation}
\label{eq:loss_det_reg}
    \mathcal{L}_{{\rm det\_reg}} = \sum_{i=0}^{N_{A}-1} \mathbbm{1}_{\{c^{i}_{A} \ne \varnothing\}} ||\hat{b}^{\hat\sigma(i)} - {b}^i||_1
\end{equation}

\paragraph{Motion Prediction.} Because PIP detects not only dynamic agents but static agents like barrier or traffic cone, and we only predict future motions for the dynamic agents, we define a dynamic agent class $C_d$ includes all agent classes in nuScenes\cite{caesar2020nuscenes} except for \textit{traffic\_cone} and \textit{barrier}. For the \textit{i}-th agent, after accumulating the predicted trajectories in offset format, we first calculate the $\textit{l}_2$ Final Displacement Error (FDE) between each predicted trajectory $\hat{s}^k_i$ and its corresponding ground-truth trajectory $s_i$: ${\rm FDE}(\hat{s}^k_i, s_i)= ||\hat{s}^k_{i, T_f}-{s}_{i, T_f}||_2$, then we select the best prediction $\hat{s}^{\hat{k}}_i$ among different modes. Finally, we calculate the motion prediction loss between the best predicted trajectory $\hat{s}^{\hat{k}}_i$ and its corresponding ground-truth $s_i$ using $\textit{l}_1$ loss:

\begin{equation}
\label{eq:loss_mot_reg}
    \mathcal{L}_{{\rm mot\_reg}} = \sum_{i=0}^{N_{A}-1} \mathbbm{1}_{\{c^{i}_{A} \in C_d\}} \sum_{j=0}^{T_f-1} ||\hat{s}^{\hat{k}}_{i,j} - {s}_{i,j}||_1
\end{equation}

The total loss is a weighted sum of the above losses:

\begin{equation}
\label{eq:loss_tot}
\begin{aligned}
    \mathcal{L} = &\lambda_1 \mathcal{L}_{{\rm det\_cls}} + \lambda_2 \mathcal{L}_{{\rm det\_reg}} + \lambda_3 \mathcal{L}_{{\rm map\_cls}} \\ &+ \lambda_4 \mathcal{L}_{{\rm map\_reg}} + \lambda_5 \mathcal{L}_{{\rm mot\_reg}}
\end{aligned}
\end{equation}

\section{Experiments}
\label{sec:experiments}
\subsection{Datasets}
\label{sec:datasets}
We evaluate PIP on the public nuScenes dataset\cite{caesar2020nuscenes}. nuScenes contains 1000 driving scenes collected in Boston and Singapore. Each scene lasts for 20 seconds and the key frames are annotated at 2Hz. The entire dataset is divided into 700/150/150 samples for train/val/test set. Over 1M images are provided by 6 surround view cameras, and there are about 1.4M bounding boxes of 23 object classes annotated in the dataset.

\begin{figure*}[]
\centering
\includegraphics[width=0.95\textwidth]{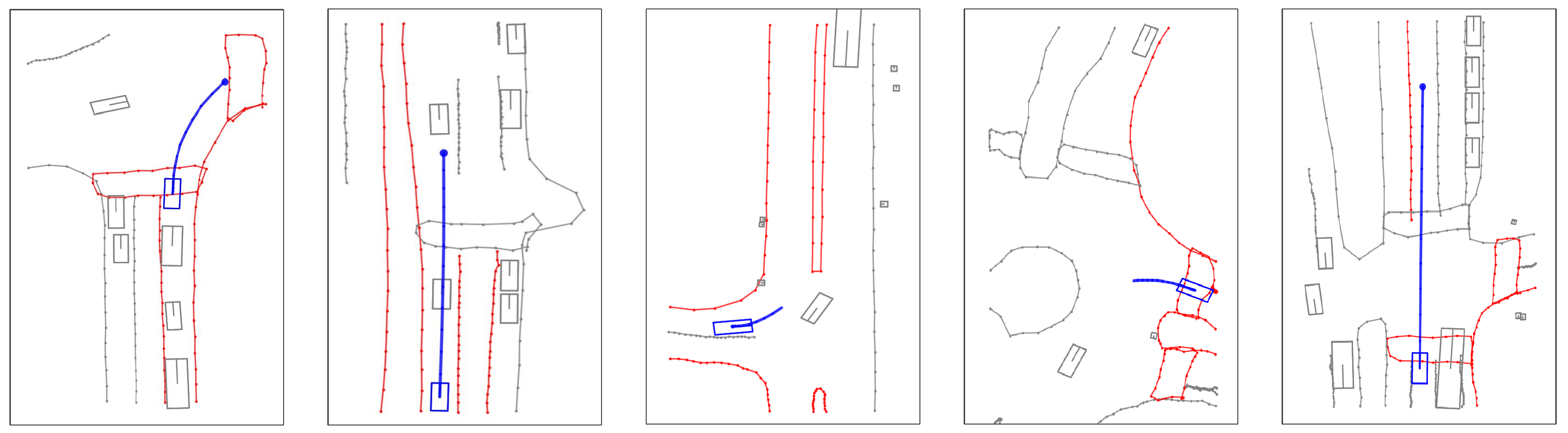} 
\caption{Visualizations of the top-4 most relevant map elements (colored red) to a target agent (denoted as blue box), according to the attention score in the cross-attention module of motion-map interaction. Blue line denotes the predicted future motion trajectory.}
\label{fig:vis_attn}
\end{figure*}

\subsection{Metrics}
\label{sec:metrics}

For evaluating the performance of motion prediction, we report three conventional motion prediction metrics: Minimum Average Displacement Error (minADE), Minimum Final Displacement Error (minFDE) and Miss Rate (MR).Considering PIP performs end-to-end motion prediction, we also report End-to-end Prediction Accuracy (EPA)~\cite{gu2022vip3d} to comprehensively and jointly evaluate detection and prediction accuracy:
\begin{equation}
\label{eq:epacost}
C_{\rm EPA}{\rm (s,\hat{s})} =
\begin{cases}
||s_0-\hat{s}_0||,& {\rm if} ||s_0-\hat{s}_0|| \leq \tau_{\rm EPA} \\
\infty,& {\rm if} ||s_0-\hat{s}_0|| > \tau_{\rm EPA}
\end{cases}
\end{equation}

\begin{equation}
\label{eq:epa}
    {\rm EPA}({S_c, \hat{S}}_c) = \frac{|\hat{S}_{\rm hit}| - \alpha N_{\rm FP}}{N_{\rm GT}}
\end{equation}
\noindent where $N_{\rm GT}$ is the number of ground-truth agent, $\hat{S}$ is the predicted agent set, and $C_{\rm EPA}$ is the matching cost between a predicted agent $\hat{s}$ and a ground-truth agent $s$, $s_0$ and $\hat{s}_0$ are the coordinates of a predicted agent and a ground-truth agent at the current timestamp. When the distance error is smaller than the threshold $\tau_{\rm EPA}$, the matching cost $C_{\rm EPA}$ between $\hat{s}$ and $s$ is equal to the distance error, otherwise it is infinity.
Next, the predicted agent which has the minimum matching cost will be defined as a matched agent for each ground-truth agent, $\hat{S}_{\rm match} \in \hat{S}$ is the matched agent set, then $N_{\rm FP} = |\hat{S}| - |S_{\rm match}|$. Finally, we use minFDE to define whether a matched agent hits its ground-truth agent, the hitted agent set is defined as: $\hat{S}_{\rm hit}=\{\hat{s}:\hat{s} \in \hat{S}_{\rm match}, {\rm minFDE}(\hat{s},s) \leq \tau_{\rm EPA}\}$. $\alpha$ is the penalty coefficient for false positive predictions.  Consistent with ViP3D\cite{gu2022vip3d}, we set $\alpha$ to 0.5, and the future time horizon is 6 second. The motion prediction metrics are computed over vehicles of all scenes.

We adopt the mean average precision (mAP) of nuScenes\cite{caesar2020nuscenes} to evaluate the 3D object detection performance. As for online mapping, We follow previous methods \cite{li2022hdmapnet, liu2022vectormapnet, liao2022maptr} which use average precision (AP) to evaluate the online mapping performance, the chamfer distance threshold is set to 1.5m to determine whether a prediction and ground-truth are matched.

\subsection{Implementation Details}
\label{sec:details}
\paragraph{Architecture.} We adopt ResNet50 pretrained on ImageNet\cite{deng2009imagenet} as the backbone network, the total historical horizon is 2 seconds for each sample, which consists of 4 key frames: \textit{t}-3, \textit{t}-2, \textit{t}-1 and \textit{t}. The spatial shapes of BEV queries are $200 \times 200$, which correspond to a $102.4{\rm m}\times102.4{\rm m}$ perception range. By default we use 300 detection queries, 100 map instance queries and 6 motion mode queries. Each map instance contains 10 points. Each encoder or decoder in the perception transformer consists of 6 layers, and the hidden size for each layer is set to 256. For agent-wise filtering, we set the confidence threshold $\tau=0.5$ and the distance threshold $\mu=20.5{\rm m}$.

\vspace{-5mm}

\paragraph{Training.} Without specific notification, we use AdamW\cite{loshchilov2017adamw} optimizer with weight decay 0.01 and cosine annealing scheduler\cite{loshchilov2016cosineanneal} with initial learning rate $2\times{10}^{-4}$.  We train all models on 8 NVIDIA GeForce RTX 3090 GPUs for 48 epochs with batch size 1 per GPU.

\subsection{Quantitative Results}
\label{sec:quantitative-results}

\paragraph{Comparison with Other Methods.}
We compare PIP with state-of-the-art methods on the nuScenes\cite{caesar2020nuscenes} val set in Table \ref{tbl:mainresults}. Without requiring human-annotated HD map, PIP outperforms these works on most of the metrics. Compared with regression-based ViP3D\cite{gu2022vip3d} which achieves state-of-the-art performance, PIP greatly reduces minADE and minFDE by 0.91m and 1.17m, as well as improving EPA and MR by 0.5\% and 5.9\%. Besides, the inference latency of PIP is 220ms on a NVIDIA GeForce RTX 3090. We will conduct more ablations about our design choices in the following sections.

\begin{figure*}[h]
\centering
\includegraphics[width=0.95\textwidth]{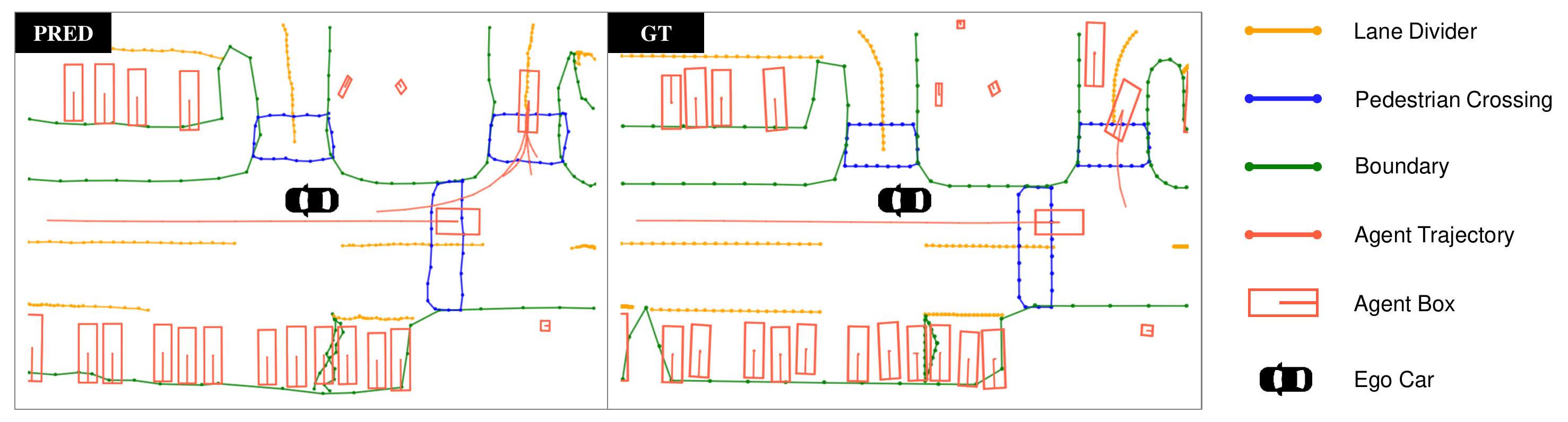} 
\caption{The qualitative results for end-to-end perception and motion prediction. PIP end-to-end outputs vectorized map, detected objects and future motion trajectories.}
\label{fig:vis_multitask}
\vspace{-2mm}
\end{figure*}

\vspace{-2mm}

\paragraph{Effectiveness of Agent-wise Normalization.}

\begin{table}
\begin{center}
\renewcommand{\tabcolsep}{3.0pt}
\begin{tabular}{c c|c c c c}
\midrule
Norm & Filter & EPA$\uparrow$ & minADE$\downarrow$ & minFDE$\downarrow$ & MR$\downarrow$ \\
\midrule
& & 0.256 & 1.31 & 1.91 & 0.197 \\
\cmark & & 0.257 & 1.28 & 1.86 & 0.196 \\
& \cmark & 0.255 & 1.29 & 1.88 & 0.195 \\
\cmark & \cmark & 0.258 & 1.23 & 1.75 & 0.195 \\
\bottomrule
\end{tabular}
\end{center}
\caption{Effectiveness of agent-wise normalization and agent-wise filtering. "Norm" means agent-wise normalization. "Filter" means agent-wise filtering.} 
\label{tbl:norm-filter}
\end{table}

As shown in Table \ref{tbl:norm-filter}, the motion prediction performance of PIP is improved on almost every metric with agent-wise normalization. This is because normalization helps to reduce the variance among agents, as well as providing agent-specific map features to guide their future motions.

\vspace{-2mm}

\paragraph{Effectiveness of Agent-wise Filtering.}

Ablations about the effectiveness of agent-wise filtering are presented in Table \ref{tbl:norm-filter}. Filtering map elements with low confidence scores and far map elements for each agent helps PIP focus on the key map elements, which provides valid prior information for agent's future motion.

\begin{table}[]
\begin{center}
\renewcommand{\tabcolsep}{3.0pt}
\begin{tabular}{c|c c c c}
\midrule
Map Interaction & EPA$\uparrow$ & minADE$\downarrow$ & minFDE$\downarrow$ & MR$\downarrow$ \\
\midrule
w/o Interaction & 0.251 & 1.30 & 1.87 & 0.201 \\
\midrule
w/ Interaction & 0.259 & 1.27 & 1.84 & 0.198 \\
+ Interaction PE & 0.258 & 1.23 & 1.75 & 0.195 \\
\midrule
HD Map & 0.262 & 1.22 & 1.72 & 0.187 \\
\bottomrule
\end{tabular}
\end{center}
\vspace{-4mm}
\caption{Effectiveness of motion-map interaction. "w/o Interaction" indicates we don't perform motion-map interaction. "HD Map" indicates using HD map instead of predicted map.} 
\label{tbl:interaction}
\end{table}

\begin{table}[]
\begin{center}
\renewcommand{\tabcolsep}{3.0pt}
\begin{tabular}{c c c|c c c}
\midrule
Map & Det & Motion & Map mAP$\uparrow$ & Det mAP$\uparrow$ & EPA$\uparrow$ \\
\midrule
\cmark & & &  0.316 & - & -  \\
& \cmark & &  - & 0.294 & - \\
\cmark & \cmark & &  0.297 & 0.280 & - \\
\cmark & \cmark & \cmark & 0.281 & 0.280 & 0.258 \\
\midrule
\end{tabular}
\end{center}
\vspace{-4mm}
\caption{Results of multi-task learning (online mapping, object detection and motion prediction) on the nuScenes\cite{caesar2020nuscenes} val set.} 
\label{tbl:multitask}
\end{table}

\vspace{-3mm}

\paragraph{Effectiveness of Motion-Map Interaction.}
Table \ref{tbl:interaction} shows the effectiveness of the motion-map interaction. With map instance queries for interaction, the motion prediction performance is improved by 0.8\%, 0.03m, 0.03m and 0.3\%in terms of EPA, minADE, minFDE and MR, respectively. With interaction PE adopted, minADE, minFDE and MR are further improved by 0.04m, 0.09m and 0.3\%, respectively. Compared with the model without motion-map interaction, the total improvement is 0.7\%, 0.07m, 0.12m and 0.6\% for EPA, minADE, minFDE and MR, respectively. We also report results using HD map instead of predicted map. The only difference is that we use map element features after agent-wise normalization and filtering as the input of VectorNet\cite{gao2020vectornet}, instead of agent-wise map queries, because there is no map queries in this setting. We can see from Table \ref{tbl:interaction} that the performance of the model using predicted map is close to the model using HD map, especially in terms of minADE, the performance gap is only 0.01m.

\paragraph{Multi-task Learning.}
Experiments about the multi-task learning of PIP are shown in Table \ref{tbl:multitask}.
In multi-task learning, tasks mutually compete. Consistent with the common sense, with more tasks included, the metric of each single task slightly drops. 

\subsection{Qualitative Results}
\label{sec:qualitative-results}

\paragraph{Visualizations of Agent-Map Correlation.}
In Figure \ref{fig:vis_attn}, we visualize the top-4 most relevant map elements to a target agent according to the attention score in the cross-attention module of motion-map intention. With the proposed interaction scheme, PIP learns scene information from the predicted vectorized map and outputs reasonable trajectories based on the road topology. 

\paragraph{Visualizations of Perception and Prediction.}
In Figure \ref{fig:vis_multitask}, we visualize the overall outputs of PIP for a complex scene (including vectorized static map, detected objects and their motion trajectories). PIP generates impressive multi-task perception and prediction results. And for motion prediction, PIP provides accurate long-straight prediction, as well as diverse multi-modal predictions for the merging vehicle. More visualizations are presented in Appendix, due to the page constraint.

\section{Conclusion}
\label{sec:conclusion}
In this work, we present PIP, an end-to-end multi-task framework based on the unified query representation. We exploit the correlation between perception and prediction with a tailored differentiable interaction scheme. 
PIP outputs high-level information (vectorized static map and  dynamic objects with motion information), and contributes to the downstream planning and control. And PIP can be extended to an interactive end-to-end perception-prediction-planning framework, which we leave as future work.


\vspace{10mm}
\appendix
\noindent{\Large \textbf{Appendix}}
\section{More Implementation Details}
We provide more implementation details about the proposed method and experiment settings in this section.
\paragraph{Map Encoding.} VectorNet\cite{gao2020vectornet} is utilized to further encode the map features, in order to achieve better motion-map interaction. In practice, it consists of three subgraph layers defined in \cite{gao2020vectornet}. Each subgraph layer contains a single MLP to encode the point-level map features, and a max-pooling layer to select the the most representative point feature within each map instance. Then this feature is concatenated with each point feature in the same map instance along the channel dimension. Finally, another max-pooling layer is used to transform point-level map features to instance-level map features.

\paragraph{Training and Evaluation.} By default, we set the multi-task loss weights $\lambda_1, \lambda_2, \lambda_3, \lambda_4$ and $\lambda_5$ to 0.8, 0.1, 0.8, 0.4 and 0.2, respectively. As for ablations, we set the corresponding loss weights to 0 when we don't perform certain tasks. During evaluation, we only consider scene samples which contain enough future frames. Besides, some agents don't have complete ground truth future trajectories because they move beyond the perception boundary. We define all these agents as "not hit" for the calculation of EPA. In terms of minADE, minFDE and MR, we only report results for agents which have complete future trajectories. After getting the predicted trajectory for each matched agent, we will translate the trajectory so that its start point is at the center of the predicted agent box. And now the predicted future trajectories become predicted future coordinates. Then we calculate minADE, minFDE between the predicted future coordinates and the ground truth future coordinates for each matched agent:

\begin{equation}
\label{eq:minade}
    {\rm minADE}(\hat{s}, s) = \underset{k \in N_{\rm mode}}{min} \frac{1}{T_f} \sum_{i=0}^{T_f-1} ||\hat{s}_i^k - s_i||_2
\end{equation}

\begin{equation}
\label{eq:minfde}
    {\rm minFDE}(\hat{s}, s) = \underset{k \in N_{\rm mode}}{min} ||\hat{s}_{T_f-1}^k - s_{T_f-1}||_2
\end{equation}

The metric results are averaged over all valid agents across different scenes. Finally MissRate (MR) can be calculated as follows:

\begin{equation}
\label{eq:mr}
    {\rm MissRate}(\hat{S}) = \frac{|\hat{S}| - |\hat{S}_{\rm hit}|}{|\hat{S}|}
\end{equation}

\noindent where $|\hat{S}|$ is the number of predicted and matched agents which have complete future trajectories across different scenes. $|\hat{S}_{\rm hit}|$ is the number of hitted agents which belong to $\hat{S}$.

\section{More Visualizations}
\begin{figure*}[h]
\centering
\includegraphics[width=0.9\linewidth]{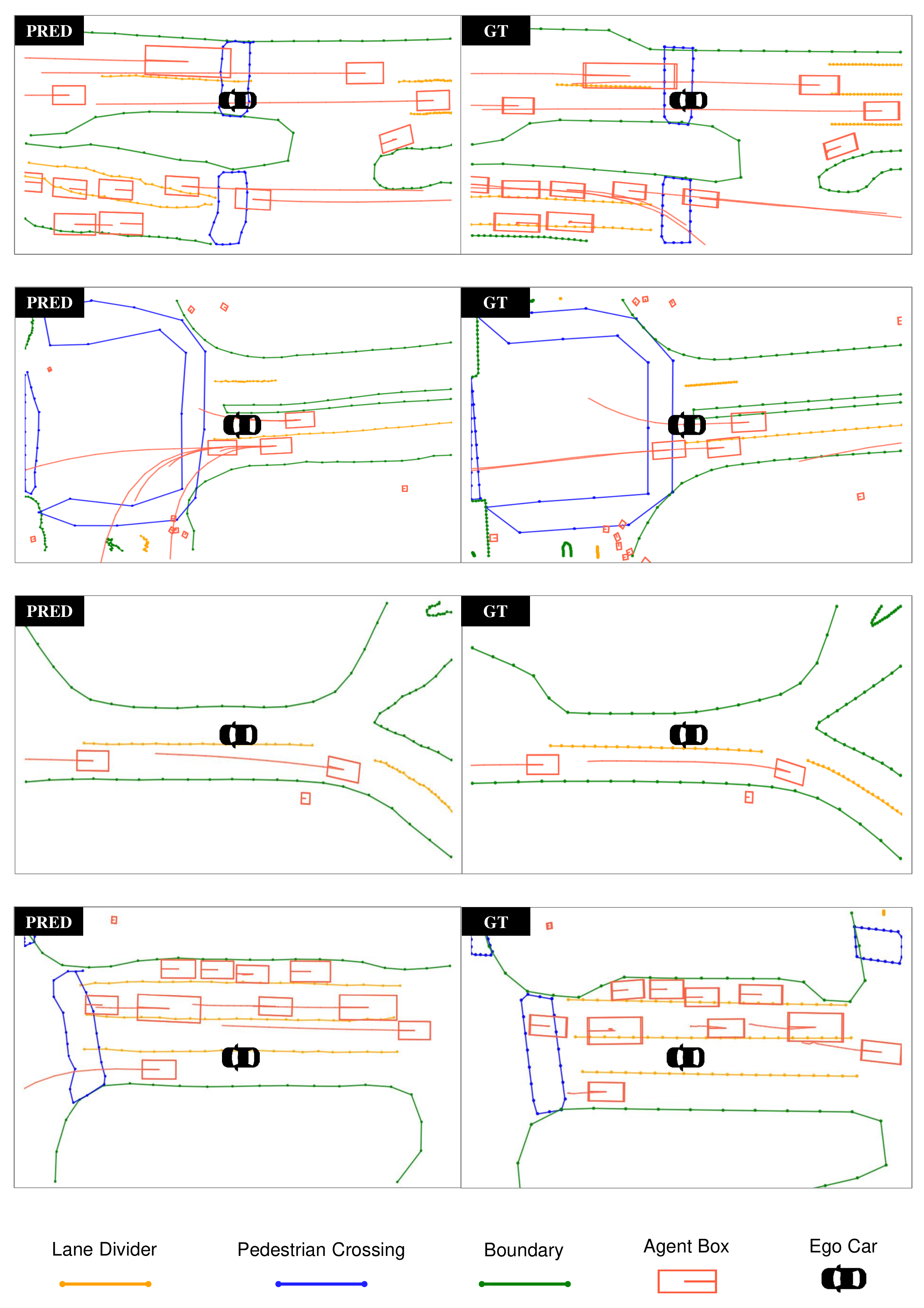} 
\caption{Qualitative results for end-to-end perception and motion prediction.}
\label{fig:supp_vis_multitask}
\end{figure*}

We show more visualizations of end-to-end perception and motion prediction in Figure \ref{fig:supp_vis_multitask}. PIP achieves accurate 3D object detection, online mapping and motion prediction performance under various complex driving scenarios.

{\small
\bibliographystyle{ieee_fullname}
\bibliography{egbib}
}

\end{document}